\documentclass[10pt,twocolumn,letterpaper]{article}
\DeclareUnicodeCharacter{0301}{\'{e}}
\usepackage{cvpr}              

\usepackage[dvipsnames]{xcolor}

\definecolor{cvprblue}{rgb}{0.21,0.49,0.74}
\usepackage[pagebackref,breaklinks,colorlinks,citecolor=cvprblue]{hyperref}

\def\paperID{*****} 
\def\confName{CVPR}
\def\confYear{2024}

\title{Rethinking Top Probability from Multi-view for Distracted Driver Behaviour Localization}

\author{Quang Vinh Nguyen$^{1}$,  Vo Hoang Thanh Son$^{1}$, Chau Truong Vinh Hoang$^{2}$, Duc Duy Nguyen$^{3}$ \\ Nhat Huy Nguyen Minh$^{2}$, Soo-Hyung Kim$^{1}$\\
$^{1}$Chonnam National University. $^{2}$Vietnamese-German University. 
\\ $^{3}$Hanoi University of Science and Technology.\\
{\tt\small \{vinhbn28,hoangsonvothanh,shkim\}@jnu.ac.kr} \\ 
{\tt\small\{16076, 10423045\}@student.vgu.edu.vn}, \tt\small duy.nd223435@sis.hust.edu.vn}

\begin{document}
\maketitle
\begin{abstract}
Naturalistic driving action localization task aims to recognize and comprehend human behaviors and actions from video data captured during real-world driving scenarios. Previous studies have shown great action localization performance by applying a recognition model followed by probability-based post-processing. Nevertheless, the probabilities provided by the recognition model frequently contain confused information causing challenge for post-processing. In this work, we adopt an action recognition model based on self-supervise learning to detect distracted activities and give potential action probabilities. Subsequently, a constraint ensemble strategy takes advantages of multi-camera views to provide robust predictions. Finally, we introduce a conditional post-processing operation to locate distracted behaviours and action temporal boundaries precisely. Experimenting on test set A2, our method obtains the sixth position on the public leaderboard of track 3 of the 2024 AI City Challenge.
\end{abstract}    
\section{Introduction}
\label{sec:intro}

Distracted driving is defined as any circumstance where the driver diverts attention away from safe driving activities. In the United States, over 3,500 lives are lost annually due to accidents caused by distracted driving. Research in intelligent transportation systems and distracted driving has gained significant attention from scholars worldwide \cite{Lv2021,Veres2020,Wang2022,Zhang2011}. This interest is fueled by the potential of naturalistic driving videos to capture real-time driving behavior and the capability of deep learning to analyze potential risk factors.
The AI City Challenge 2024 \cite{aicity2024} aims to advance research in this field by hosting a naturalistic driving action recognition challenge. The given challenge focuses on detecting distracted driving behaviors using synthetic naturalistic data collected from three camera locations inside the vehicle. This challenge involves analyzing synchronized video recordings from drivers engaged in various distracted driving activities. These activities are classified into different actions, such as using a phone, eating, and reaching into the backseat, each of which can potentially lead to accidents. 

Previous studies \cite{zhou2023, Le2023, Li2023, Tran2022, Zhao2022} have demonstrated the effectiveness in distracted driving detection, typically dividing the task into two main stages: activity recognition and temporal action localization. However, several challenges remain: (1) The dataset is limited to 16 behavior categories, leading to an insufficient diversity of samples within each category. (2) The models must discern various actions from different perspectives within untrimmed videos, facing difficulties in distinguishing subtle variations within the same class and detecting minor discrepancies between certain classes. (3) The inclusion of the appearance block constrains the model's ability to discern differences between certain classes. (4) Previous solutions rely heavily on the classification model's confidence, which can result in misclassifications when the highest and second-highest classes have similar probabilities.

Therefore, in this paper, we aim to contribute to the literature in the following manners: First, we inherit an action classification model in video based self-supervised learning to detect robust distracted actions from the input video. Next, we apply a constraint ensemble strategy to take advantage of the power of each camera view. In the final, conditional post-processing steps consider contexts from top 1 and top 2 confidence ranking to locate distracted actions and temporal boundaries accurately.

\section{Related Work}
\label{sec:formatting}


\subsection{Action Recognition}

Action recognition is a crucial task in the field of video understanding. Over the years, there have been numerous studies and extensive research conducted in this area. The main goal of the action recognition is to classify a trimmed video into specific action classes using end-to-end deep learning methods. There have been significant updates in architecture design, ranging from 2D-based CNN models and 3D-based CNN models to Transformer-based models.

2D-based action recognition methods first implement a CNN model to extract spatial features for each frame in the video. The sequence models\cite{Wang2016, Donahue2015} are employed to fuse these features with the aim of capturing temporal information. 3D-CNN attempts\cite{Carreira2017,Feichtenhofer2019, Tran2015} to process spatial-temporal information directly by using 3D input tensors, where 2 dimensions represent space and 1 dimension represents time . The success of Transformer in image-related and sequential tasks and has motivated the exploration of its potential in video recognition, \cite{Li_Wu_Fan_Mangalam_Xiong_Malik_Feichtenhofer_2022, Liu_Ning_2022} have been successfully developed to use Transformer in the architecture. Recent works also take advantages of large video foundation pre-training models to improve performance. Masking with high ratio or scaling transformer model by applying self-supervised learning,  \cite{videomae, Wang_Huang2023, Wang_Li_2022}  have shown great potential in extracting robust video representation

\subsection{Temporal Action Localization}

Temporal action localization is the task of automatically identifying the time duration during which an action occurs within an untrimmed video and determining its corresponding action category. The conventional two-stage method involves proposing action segments initially and subsequently classifying these proposals into their respective action categories \cite{Xu2017,Lin2019,Lin_Gan_Han2019,Qing2021}. However, a major drawback of this method is that the boundaries of action instances remain fixed during the classification process. As a result, while the method can identify time intervals likely to contain actions, it lacks the ability to precisely determine the exact start and end times of the actions.

In contrast, one-stage methods have garnered significant attention by integrating the localization and classification tasks within a single network. This approach eliminates the fixed boundaries issue and offers a more streamlined solution. Previous works have seen the adoption of hierarchical architectures based on CNN \cite{Zhao_Xiong_2019, Lin_Gan_Han2019, Li_Ji_2020}. Recent studies \cite{Avidan2022, Shi2023} extract a video representation with a Transformer-based encoder.

\section{Method}
\label{sec:method}
\begin{figure*}[htbp]
  \centering

   \includegraphics[width=1\linewidth]{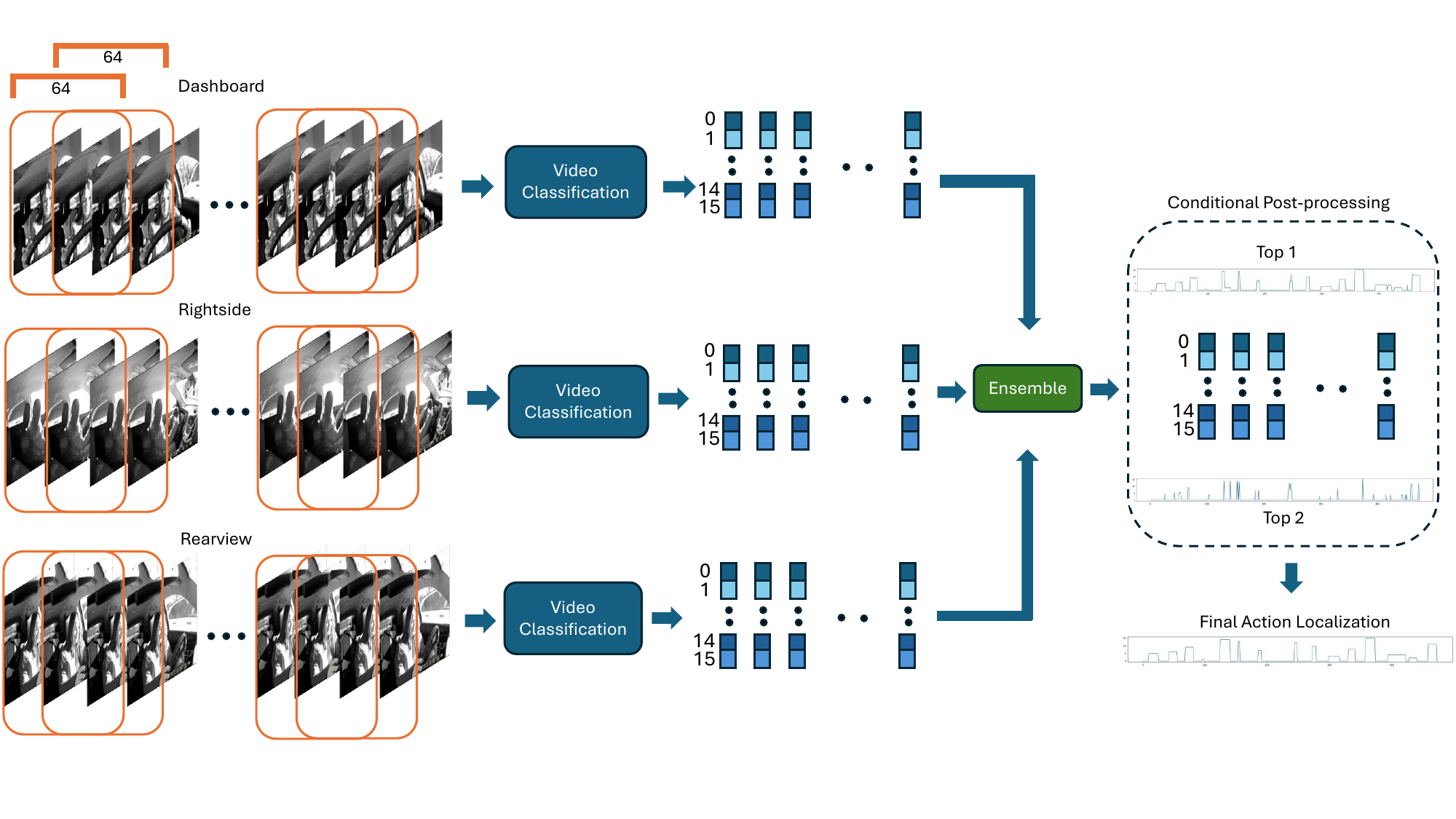}

   \caption{Distracted Driver Behaviour Recognition System}
   \label{fig:model}
\end{figure*}

As indicated in \cref{fig:model}, our distracted driver behaviour recognition system consists of three main novel components: an action recognition model, an ensemble strategy, and conditional post-processing. The first is an action recognition model which is self-supervised learning, recognize distracted driver behaviors from input short videos. The second is an ensemble strategy being responsible for integrating multi-view predictions. Given recognition probabilities, conditional post-processing considers diverse contexts to smooth out detected activities and localize the temporal boundary accurately. Detailed descriptions of each component are presented in the following subsections.

\subsection{Action Recognition}

Recent researches have demonstrated that self-supervised learning (SSL) can provide more robust \cite{Dan} and general features \cite{Kim2021, Jean,Xinlei}, while reducing the amount of data required for an equivalent supervision-based pre-training. In the context of video understanding, self-supervised learning techniques seek to take advantage of the temporal coherence and spatial correlations seen in video sequences. These approaches are particularly suitable for scenarios where labeled data is scarce or expensive to obtain as the distracted driver behavior dataset. Inspired by the successful study of Masking Modeling in the text and picture domain \cite{Devlin,He,Chen}. VideoMAE \cite{videomae} employs Masked Autoencoders, a variation on traditional Autoencoders where certain parts of the input data are masked out during training, encouraging the model to learn useful representations that capture the underlying structure of the data leading to promising performance in a variety of video understanding tasks. Our system inherits this structure to classify distracted action from naturalistic driving videos. Specifically, input videos with FPS 30 are trimmed into a series of short videos containing 64 frames. The model achieves short videos as input to give the probability for each class in the output. 

\subsection{Multi-view Ensemble Strategy}
\begin{figure}[t]
  \centering

   \includegraphics[width=1.7\linewidth]{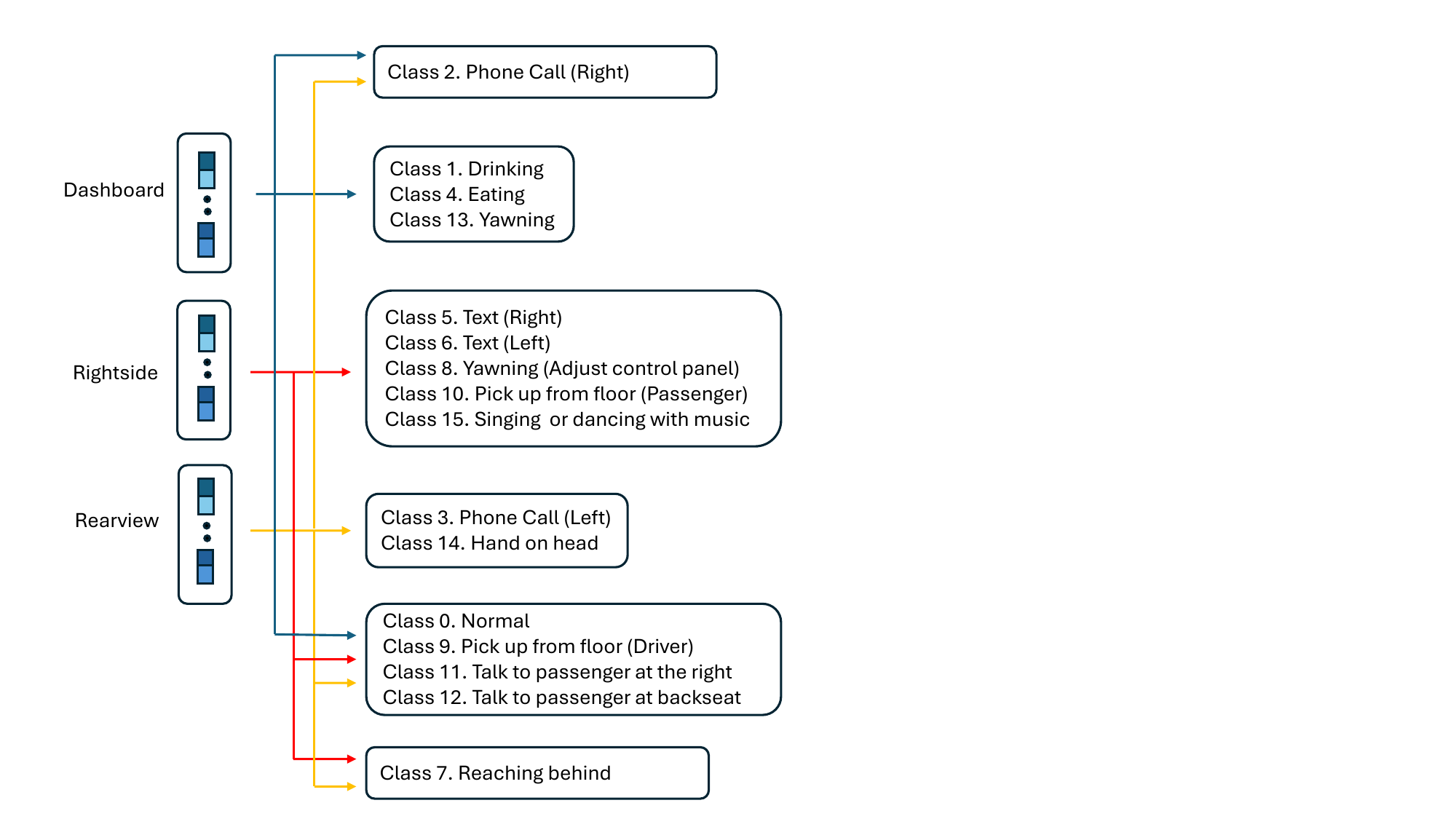}

   \caption{Ensemble strategy}
   \label{fig:ensemble}
\end{figure}

The distracted driver action is divided into sixteen distracted actions and three views of the camera mounted in the car: dashboard, rearview and rightside. Each of these views has significance in different contexts. Dashboard view directly facing driver contributes clearly to actions: "phone call by right hand", "drink", "eating" or activities involving to the movement of body-head such as "talk with passenger", "pick up from floor". Rear view gives a broader space view inside the car, and is useful for identifying various actions: "phone call by right hand or left hand", "reaching behind" or "hand on head". While the right side view shows a different view, from the right side of the driver, this view is helpful for hand movements: "control the panel", "text by hand" or "pick from floor (Passenger)". In addition, several specific classes: "talk with passenger", "pick from floor (Driver)" can be integrated by all views to comprehend the overall context of distracted driving. Therefore, in order to enhance recognition performance, we suggest an ensemble strategy based on multi-view. The specifics of ensemble strategy are displayed in \cref{fig:ensemble}. 

\subsection{Conditional Post-Processing}
The action recognition model classifies short videos which are trimmed from input video to give a series of probability. Output probability is an array of prediction vectors with the length of 16 corresponding to a number of classes. Elements with the highest value in vectors refer to predicted classes. And elements with second highest value normally express potential classes which are the second most trustworthy after the highest ones. Our post-processing strategy leverages top 1 and top 2 of output probability to locate the actions and time boundary more accurately. This process consists of three main steps: Conditional Merging, Conditional Decision and Missing Labels Restoring. 

\begin{figure}[h]
  \centering
   \includegraphics[width=1\linewidth]{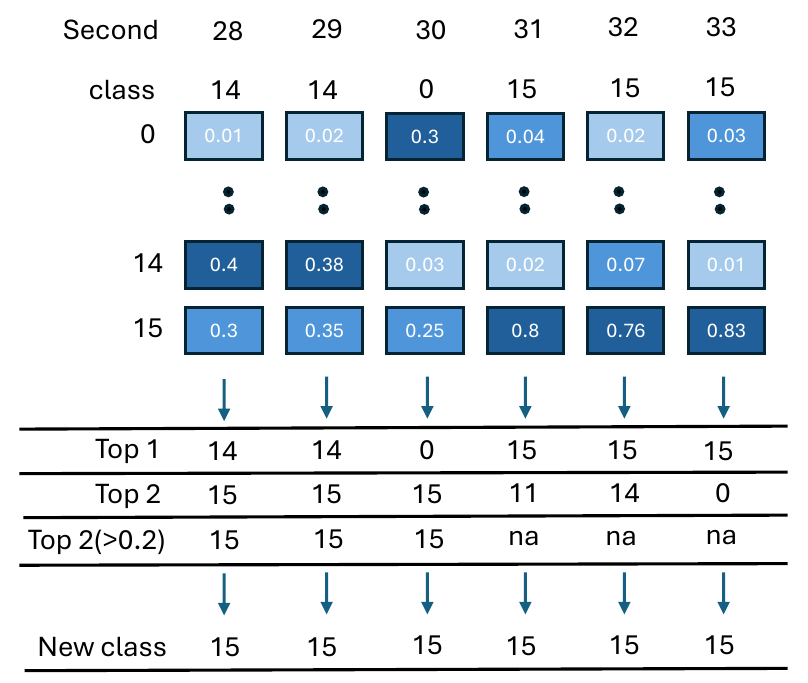}
   \caption{Conditional Merging}
   \label{fig:merge}
\end{figure}

\noindent\textbf{Conditional Merging.} The first operation refers to conditional merging, which is depicted in \cref{fig:merge}. Instead of merging closer actions normally, this component considers the context of one certain class and neighbor classes by top 1 and top 2 confidence ranking to merge potential candidates and remove noise classes. To explain symbols in \cref{fig:merge}, "second" represents the time boundary for each action, the values in the boxes refer to the probabilities for each type. Top 1 is the class with the highest probability score, while top 2 represents the class with the second highest probability.

\begin{figure}[t]
  \centering
  
   \includegraphics[width=1\linewidth]{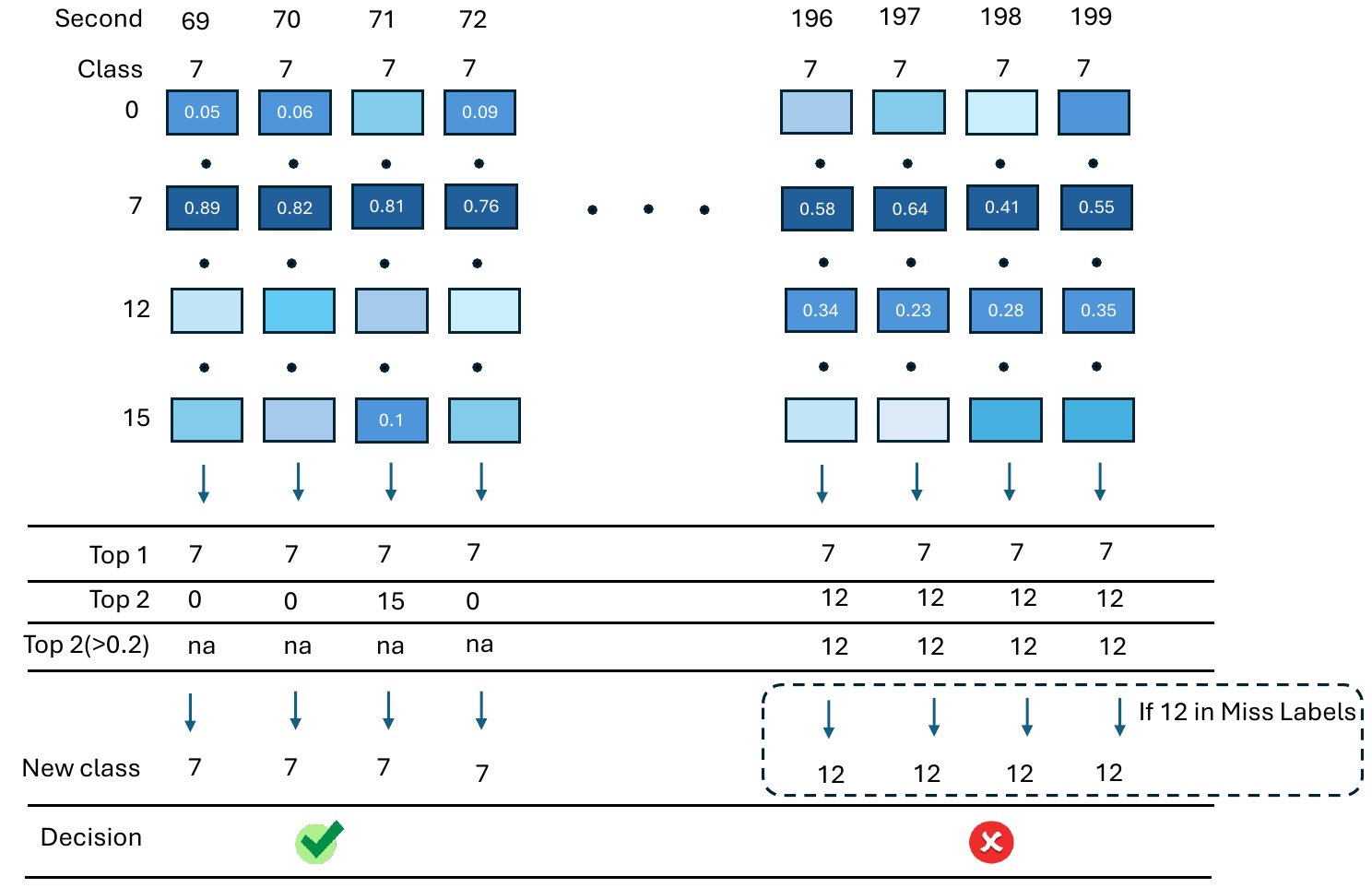}

   \caption{Conditional Decision}
   \label{fig:decision}
\end{figure}

\noindent\textbf{Conditional Decision.} \cref{fig:decision} describes the conditional decision operation which selects a reliable time segmentation from different segments of the same classes. Given several different segments of the same class, for example, there are two segments of class 7 "reaching behind" in \cref{fig:decision}. The decision module relies on probabilities from top 1 and top 2 to filter a most trustworthy segment. 

\noindent\textbf{Missing Labels Restoring.} After the two mentioned above steps, it still has some classes that are missing or not detected by the top 1 prediction. It means that if we just use top 1 probability for output prediction, the system could not localize distracted actions sufficiently. The restoring module shown in \cref{{fig:restore}} finds these classes to reproduce the final prediction with enough 16 classes.

\begin{figure}[h]
  \centering

   \includegraphics[width=1.\linewidth]{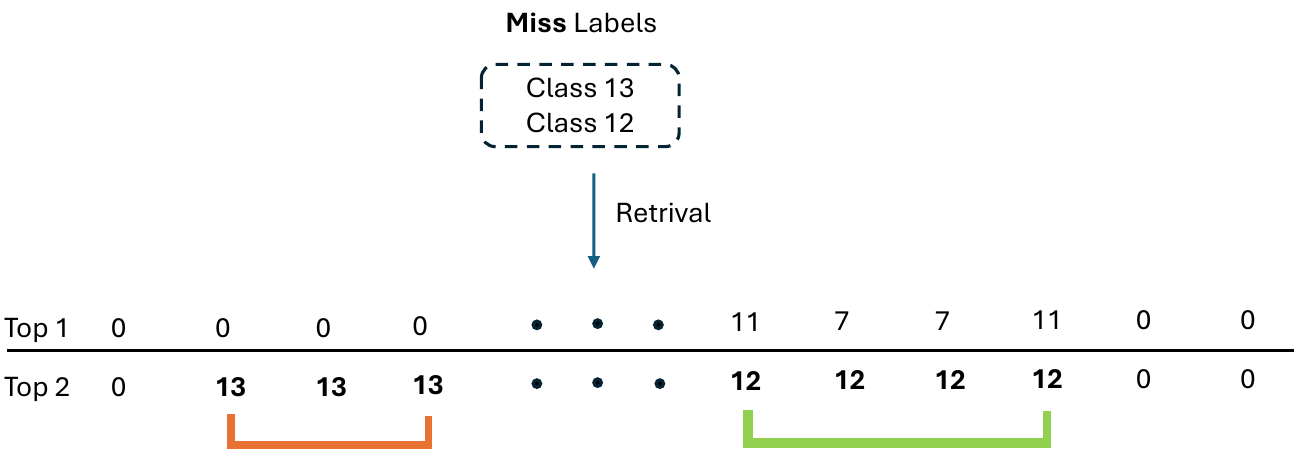}

   \caption{Missing Labels Restoring}
   \label{fig:restore}
\end{figure}

\section{Experiments}
\subsection{Dataset}

\begin{table*}[htbp]
\footnotesize	
\centering
\begin{tabular}{c|c|c|c |c| c| c| c| c| c| c| c| c| c| c| c|c|c}
 Fold& View& 1 & 2 & 3 & 4 & 5 & 6 & 7 & 8 & 9 & 10 & 11 & 12 & 13 & 14 & 15&Avg \\ \hline
 &  Dash & 1.00 & 0.94 & 0.94 & 0.80  & 0.90 & 0.98 & 0.89 & 0.97 & 0.75 & 0.81 & 0.84 & 0.86 & 0.75 & 0.95 & 0.81 & 0.88\\
 1&   Rear & 1.00 & 0.96 & 0.99 & 0.81 & 1.00 & 0.91 & 0.96 & 0.94 & 0.84 & 0.89 & 0.70 & 0.82 & 0.82 & 0.96& 0.80 & 0.89\\
 &   Right & 1.00 & 0.99 & 0.96 & 0.66 & 0.94 & 0.92 & 0.86 & 0.96 & 0.86 & 0.81 & 0.85 & 0.70 & 0.80 & 0.90&0.80 &0.88 \\
 &   \textbf{Ensemble} & 1.00 & 0.96 & 0.99 & 0.80 & 0.94 & 0.92 & 0.97 & 0.96 & 0.82  & 0.81 & 0.86 & 0.88 & 0.75 & 0.96 & 0.80 & \textbf{0.90} \\ \hline
 &   Dash & 1.00 & 0.88 & 0.98 & 0.70 & 0.96 & 0.95 & 0.90 & 0.86 & 0.76 & 0.66 & 0.76 & 0.68 & 0.61 & 0.96& 0.88 & 0.84 \\
 2&   Rear & 1.00 & 0.94 & 0.97 & 0.67 & 0.96 &0.96  &0.95  & 0.83 & 0.80 &  0.75& 0.64 &0.74  &0.60  &0.98 & 0.84 &0.84\\
 &   Right & 0.96 & 0.85 & 0.91 & 0.27 & 0.96 & 1.00 & 0.93 & 0.99 & 0.81 &0.75  &0.74  & 0.58 & 0.42 & 0.84& 0.82 &0.79\\
 &    \textbf{Ensemble} & 1.00 & 0.95 & 0.97 & 0.70 & 0.96 & 1.00 & 0.95 & 0.99 & 0.80 & 0.75 & 0.70 & 0.72 & 0.61 & 0.98 & 0.82 & \textbf{0.86} \\ \hline
 &   Dash & 1.00 & 0.98 & 0.68 & 0.80 & 0.88 & 0.88 & 0.88 & 0.93 &0.78  & 0.61 & 0.60 & 0.68 & 0.87 & 0.97& 0.77 &0.82\\
 3&   Rear & 1.00 & 1.00 & 0.63 & 0.83 & 0.94 & 0.97 & 0.86 & 0.98 &0.85  & 0.76 & 0.69 & 0.71 & 0.81 & 0.97& 0.75 & \textbf{0.85}\\
 &   Right & 0.88 & 0.99 & 0.60 & 0.75 & 0.99 & 0.92 & 0.88 & 0.98 & 0.92 & 0.76 & 0.72 & 0.67 & 0.63 & 0.97 & 0.76 & 0.83  \\
 &   \textbf{Ensemble} & 1.00 & 1.00 & 0.63 & 0.80 & 0.99 & 0.90 & 0.88 & 0.98 & 0.84 & 0.75 & 0.64 & 0.70 & 0.87 &0.97 & 0.76 & \textbf{0.85}\\ \hline
   &   Dash & 0.92 & 0.97 & 0.97 & 0.81 & 0.96 & 0.90 & 0.85 & 0.88 & 0.76 & 0.17 &  0.86& 0.68 &0.86  &0.93 &0.85 &0.78 \\
4 &   Rear & 0.89 & 0.96 & 0.99 &0.81  & 0.92 & 0.80 & 0.82 & 0.85 & 0.73 & 0.15 & 0.86 & 0.65 & 0.84 & 0.94& 0.86 &0.80\\
 &   Right &  0.94& 0.97 & 0.90 & 0.67 & 0.97 & 0.92 & 0.91 & 0.97 &0.86  & 0.20 & 0.77 &0.66  &0.60  & 0.95& 0.83 &0.81\\
 &   \textbf{Ensemble} & 0.92 & 0.97 & 0.99 & 0.96 & 0.97 & 0.92 & 0.90 & 0.97 & 0.88 & 0.20 & 0.86 & 0.66 & 0.86 & 0.94& 0.83 & \textbf{0.86}  \\ \hline
  &   Dash & 0.94 &0.91  &0.95  &0.79  &0.87  & 0.84 & 0.78 & 0.78 & 0.85 & 0.79 & 0.80 & 0.69 & 0.82 & 0.88& 0.80 &0.83 \\
 5&   Rear & 0.87 & 0.96 & 0.93 & 0.66 & 0.96 & 0.94 & 0.93 & 0.93 & 0.85 &0.79  &0.82  &0.68  & 0.92 &0.97 & 0.84 & 0.87\\
 &   Right & 0.94 & 0.98 & 0.88 & 0.46& 0.88 & 0.96& 0.91 & 0.97 & 0.89 & 0.88 & 0.56 & 0.77 & 0.73 & 0.95 & 0.84 & 0.84 \\
 &   \textbf{Ensemble} & 0.94 & 0.95 & 0.93 & 0.79 & 0.88 & 0.96 & 0.96 & 0.97  & 0.86  & 0.90 & 0.82 &  0.70 & 0.82 & 0.97 & 0.84 & \textbf{0.89} \\ \hline
\end{tabular}
\caption{The accuracy on the validation set of each 5-Fold split in different classes.}
\label{tab:accuracy}
\end{table*}

The distracted driver behavior dataset provides a comprehensive collection of driving videos capturing the actions of 99 individual drivers over a total of 90 hours. Each driver is recorded performing a series of 16 different distracting activities randomly, with the order of these activities also randomized within each video. To ensure a holistic view of the driving scenario, the dataset employs three cameras simultaneously recording from different angles within the car. Notably, each driver undergoes two rounds of data collection: one without any form of distraction and another with a predetermined distractor, such as sunglasses or a hat. This design allows for a thorough examination of driver behavior under varying levels of distraction, offering valuable insights into the impact of external factors on driving performance. The videos from the 2024 AI City Challenge's Track 3 on Naturalistic Driving Action Recognition are separated into two datasets: "A1" for training, "A2" for testing with the training dataset "A1" containing the ground truth labels for the start time, end time, and types of distracted actions.

\subsection{Evaluation Matric}

\begin{figure*}[ht!]
  \centering

   \includegraphics[width=1\linewidth]{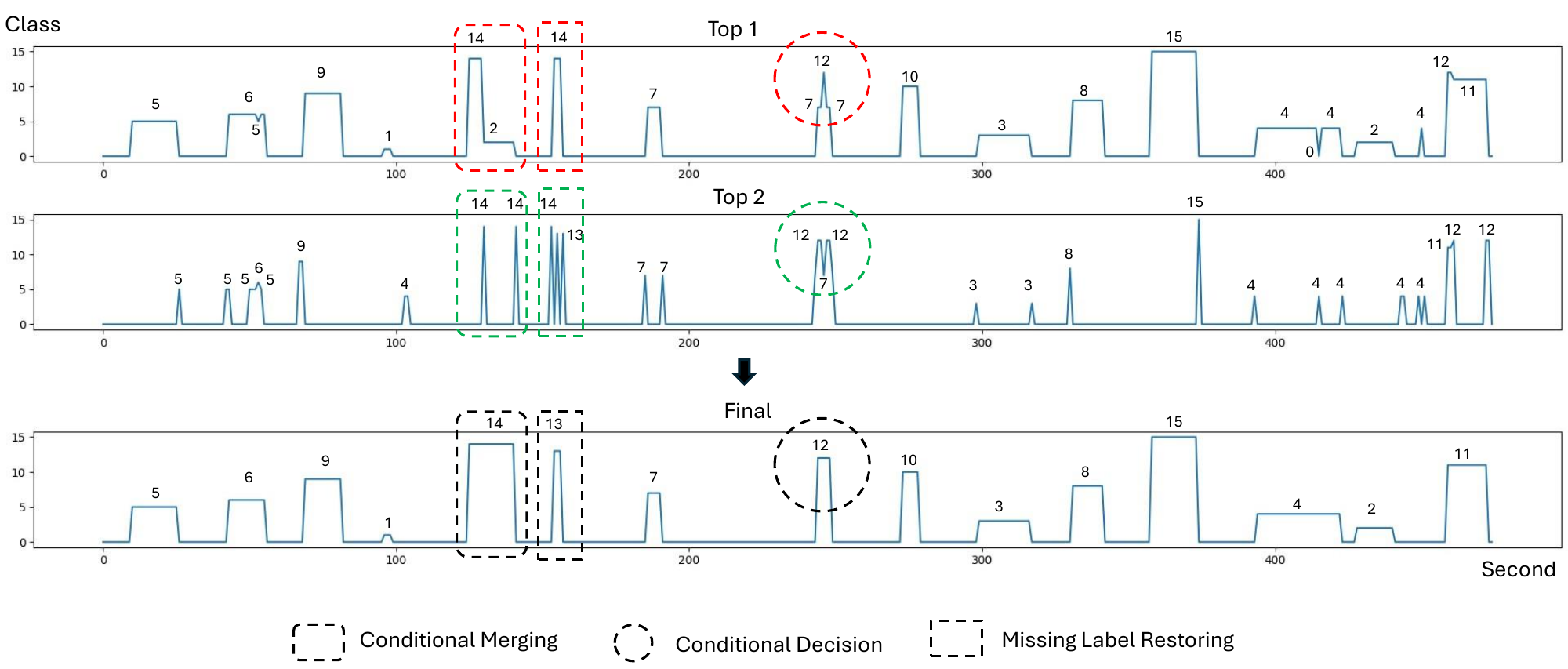}

   \caption{Action localization result of Conditional Post-Processing}
   \label{fig:result}
\end{figure*}

\textbf{Action Recognition.} Action classification involves the task of assigning a label or category to a video based on its content. The accuracy score is calculated by comparing the predicted class labels with the ground truth labels for all videos. A higher accuracy score indicates better performance of the video classification model in correctly predicting the class labels of videos. The accuracy is defined as:
\begin{equation}
Accuracy=\frac{\textit{Number of Correct Predictions}}{\textit{Total Number of Predictions}} \textit{X} 100\%
\end{equation}

"Number of Correct Predictions" is the number of instances that are correctly classified by the classifier. "Total Number of Predictions" is the total number of instances in the dataset.

\noindent\textbf{Temporal Action Localization.} For temporal action localization, activity overlap measure (os) quantifies the degree of overlap between the predicted temporal segment and the ground truth annotation for a particular action or activity within a video sequence. 

\begin{equation}
os=\frac{Intersection}{Union}
\end{equation}

Intersection is the duration of time that is common to both the predicted segment and the ground truth annotation. Union is the total amount of time covered by both the predicted segment and the ground truth annotation.

\subsection{Implement Detail}
The methodology employed relies on the PyTorch framework, a publicly available toolbox widely used in machine learning research. All experimentation was conducted on a high-performance workstation equipped with two RTX 3090 graphics card boasting 48GB of memory. For the video classification task, the network architecture utilized is consistent with the model described in reference \cite{videomae}. In particular, we use a standard Vision Transformer (ViT) model as the foundation. Each input video trimmed with stride 30 frames comprises 64 frames, sampled 16 frames evenly spaced per video. Training process is conducted with a learning rate of 2 x 10-3 over 20 epochs for each camera view.

\subsection{Results}
\textbf{Action Recognition.}
The training dataset A1 is divided into 5 folds. We validate each of the folds in all three views of the camera. Results in \cref{tab:accuracy} illustrate the effect of each of views on different classes. As can be seen, the right side view often gives excellent accuracy in several classes such as class 8 (control the panel), class 10 (pick up from floor of passenger), or class 5,6 (text) because this view is expert in these classes more than rear view and dashboard view. Besides, the dashboard view contributes greatly to class 1 (drink), class 4(eat), or class 13(yawning) and often is the best. In addition, the rear view strongly affects performance of class 3 (phone call by left hand), and class 14(hand on head). Our ensemble strategy improves and surpasses situations with only a single view. Results in each of the folds fluctuate and depend on the challenge of the validation set. 

\begin{table}[b]
  \centering
  \begin{tabular}{@{}lc@{}c@{}}
    \toprule
    Rank & Team ID & Score \\
    \midrule
    1 & 155 & 0.8282 \\
    2 & 189 & 0.8213 \\
    3 & 32 & 0.8149\\
    4 & 207 & 0.8045 \\
    5 & 5 & 0.7798 \\
    6 & \textbf{136} & 0.7625\\
    7 & 17 & 0.6844 \\
    8 & 165 & 0.6080 \\
    9 & 156 & 0.5963\\
    10 & 125 & 0.2307\\
    \bottomrule
  \end{tabular}
  \caption{Leaderboard of challenge track.}
  \label{tab:example}
\end{table}

\noindent\textbf{Temporal Action Localization.} 
The proposed method is trained on the A1 dataset provided by the competition, and tested on public test dataset A2 to evaluate temporal action localization performance. As indicated in \cref{tab:example}, our approach ranks 6th on the leaderboard with a 0.76 os score, outperforms 7th by almost 8\% score and is far ahead of competitors beneath. Besides, our solution is not much lower than the top-rank methods. This proves the effectiveness and potential of introduced method in the distracted driver behaviour recognition challenge. \cref{fig:result} depicts post-processing operation in detail, Horizontal axis denotes for time variable (second), and vertical axis refers to classes (from 0 to 15). Numbers on top of bars in the \cref{fig:result} express corresponding classes. The top 1 chart shows prediction given by highest confidence probability, while the top 2 illustrates second reliable classes. As can be seen in the top 1 chart, the predicted labels attach with many noisy labels causing confusion to action recognition and localization. The proposed post-process operation considers the top 1, top 2 probabilities, applies conditional merging, conditional decision and missing label restoring to smooth and localize accurately distracted action prediction. \cref{fig:result} indicates that our final result is seamless and superior to the top 1 prediction. This demonstrates that our post-processing strategy help model make decisions accurately and effectively localize temporal boundaries.

\section{Conclusion}

In this work, we have suggested a conditional recognition system for the distracted driver behaviour localization task. First, our method uses a pre-trained action recognition model that was trained by self-supervised learning to identify distracted activities in video input. After that, a multi-view ensemble strategy is adopted to leverage the advantages of each camera view. Given output probabilities, we post-processing by conditional merging, conditional decision, and missing labels restoring operation to recognize the distracted actions and locate time boundary accurately. Consequently, we achieved the sixth rank score in test set "A2", surpassing methods ranked lower while remaining very close to the top ranking.

\section{Acknowledgement}
This work was supported by the National Research Foundation of Korea(NRF) grant funded by the Korea government(MSIT) (RS- 2023-00219107). This work also was supported by Institute of Information \& communications Technology Planning \& Evaluation (IITP) under the Artificial Intelligence Convergence Innovation Human Resources Development (IITP-2023-RS-2023-00256629) grant funded by the Korea government(MSIT)"
{
    \small
    \bibliographystyle{ieeenat_fullname}
    \bibliography{main}
}


\end{document}